\pdfoutput=1
\documentclass[11pt]{article}

\usepackage[preprint]{acl}
\usepackage{pdfpages}
\usepackage{subcaption}
\usepackage{balance}
\usepackage{float}

\usepackage{times}
\usepackage{latexsym}
\usepackage{multirow}
\usepackage{booktabs} 
\usepackage[most]{tcolorbox}
\usepackage{listings}
\usepackage[T1]{fontenc}

\usepackage[utf8]{inputenc}

\usepackage{microtype}

\usepackage{inconsolata}

\usepackage{graphicx}

\title{A Novel Framework for Automated Explain Vision Model Using Vision-Language Models}

\author{
 \textbf{Phu-Vinh Nguyen$^{1, *}$},
 \textbf{Tan-Hanh Pham$^{2, *}$},
 \textbf{Chris Ngo$^{3}$},
 \textbf{Truong Son Hy$^{4}$}
\\
\\
 \textsuperscript{1}Uppsala University, Sweden,\\
 \textsuperscript{2}Harvard University, USA,
 \textsuperscript{3}Knovel Engineering Lab, Singapore, \\
 \textsuperscript{4}Department of Computer Science, University of Alabama at Birmingham, USA \\
\\
 \small{
   $^{*}$Equal contribution} \\
  \small{
   \textbf{Correspondence:} \href{hanhpt.phamtan@gmail.com}{hanhpt.phamtan@gmail.com}
 }
}

\begin{document}
\maketitle
\begin{abstract}
The development of many vision models mainly focuses on improving their performance using metrics such as accuracy, IoU, and mAP, with less attention to explainability due to the complexity of applying xAI methods to provide a meaningful explanation of trained models. Although many existing xAI methods aim to explain vision models sample-by-sample, methods explaining the general behavior of vision models, which can only be captured after running on a large dataset, are still underexplored. Furthermore, understanding the behavior of vision models on general images can be very important to prevent biased judgments and help identify the model's trends and patterns. With the application of Vision-Language Models, this paper proposes a pipeline to explain vision models at both the sample and dataset levels. The proposed pipeline can be used to discover failure cases and gain insights into vision models with minimal effort, thereby integrating vision model development with xAI analysis to advance image analysis. For more details, please visit the project GitHub repository: \url{https://github.com/phuvinhnguyen/autoXplain}.

\end{abstract}

\section{Introduction}
Understanding how vision models make decisions is important to improve the reliability and trustworthiness of AI systems. Although there are many established methods, benchmarks for evaluating the overall performance of vision models on large datasets, methods focusing on analyzing how models understand images, especially on large image datasets, are still limited despite the importance of explainability in providing information about how and why the model fails in some scenarios. Consequently, a scalable pipeline to explain vision models in one sample or a large vision dataset would be important for image processing development.

xAI methods such as CAM, GradCAM, ScoreCAM, LIME, and TCAV are introduced to explain vision models. Despite most of the xAI methods focusing on the instance-level, where they try to explain one sample at a time, some methods like TCAV focus on the dataset-level. However, TCAV depends on the quality of the provided concept, which can be difficult to gather completely if we have too many images. An existing framework, LangXAI, uses VLM to generate a description of how vision models attend to an image. Although the process is completely automatic, using this method on a large dataset requires manually summarizing and analyzing the generated descriptions to understand how the model generally performs.

This work includes three main contributions. First, we propose a scalable pipeline that combines CAM-based methods with VLMs to explain the behavior of the vision model. Second, we propose masked CAM images, which show the benefit of understanding the attended regions of vision models in this study's scope. Lastly, we introduce a confusion matrix used in the pipeline, which helps summarize models' behavior on a large dataset, providing a general understanding of the models.

\section{Related Work}

Although many frameworks focus on evaluating vision model performance with metrics like accuracy, IoU, ensuring transparency and interpretability through explainable AI (xAI) is also crucial~\cite{gunning2019darpa, zhao2015saliency}. xAI includes a variety of techniques to make machine learning models more interpretable and is generally classified as model-agnostic and model-specific methods~\cite{lundberg2017unified}. Model-agnostic approaches, applicable to any model, often assess feature importance, while model-specific methods leverage internal model structures for explanation~\cite{bach2015pixel}. For vision tasks, popular techniques such as LIME~\cite{ribeiro2016should}, TCAV~\cite{kim2018interpretability}, and CAM-based methods, including CAM~\cite{CAM}, Grad-CAM~\cite{selvaraju2017grad}, Grad-CAM++\cite{Chattopadhyay2017GradCAMGG}, LayerCAM~\cite{Layercam}, ScoreCAM~\cite{ScoreCAM}, EigenCAM~\cite{muhammad2020eigen}, and XGradCAM~\cite{Oquab_2015_CVPR, wang2020score} highlight regions important for predictions~\cite{itti1998model, kummerer2014deep, zhao2015saliency}. These tools are especially valuable in fields like healthcare~\cite{borys2023explainable, kakogeorgiou2021evaluating, kim2022xai}, although many still require expert interpretation, which poses challenges to integration into development workflows.

The development of Vision-Language Models (VLMs) expands the capabilities of LLMs such as Qwen~\cite{Bai2023QwenTR}, Llama~\cite{Touvron2023LLaMAOA}, and Phi~\cite{Li2023TextbooksAA} by enabling them to process images and text simultaneously~\cite{ranasinghe2024learning, NEURIPS2023_6dcf277e}. VLMs use vision models such as CLIP~\cite{radford2021learning} to excel in multimodal tasks. Prominent examples include Flamingo~\cite{alayrac2022flamingo}, BLIP~\cite{li2022blip} integrates a visual encoder with an LLM via a querying transformer~\cite{li2023blip}, and different VLMs such as GPT-4o, Qwen-VL~\cite{Bai2023QwenTR}, and Llama Vision~\cite{chu2024visionllama}, show strong ability to understand visual data. Consequently, they are widely used in many applications, including evaluating existing vision models~\cite{judgemllm}.

Despite the importance of xAI and the significant advancement in VLMs in recent years, the applications to analyze interpretive visualizations, such as Grad-CAM, in visual models remain underexplored. To fill this gap, LangXAI~\citet{nguyen2024langxai} explored the potential of using VLMs to generate explanations for visual recognition based on the intensity of colors extracted from CAM methods. However, the framework generates a description for one sample at a time without summarizing, evaluating, and comparing the general interpretability of models on a set of images, making it difficult to understand their general underlying features and behaviors, as we cannot just read many descriptions for each model. To further bridge this gap, we developed a scalable pipeline that utilizes VLMs to evaluate predictions from vision models, scoring them, providing detailed explanations, and summarizing the model's attention with a confusion matrix on a larger dataset. This method overcomes previous work by providing quantitative results on a larger dataset, helping to generalize the use of xAI and better connect training with understanding.

\section{Methodology}

\begin{figure*}
    \centering
    \includegraphics[width=0.9\linewidth]{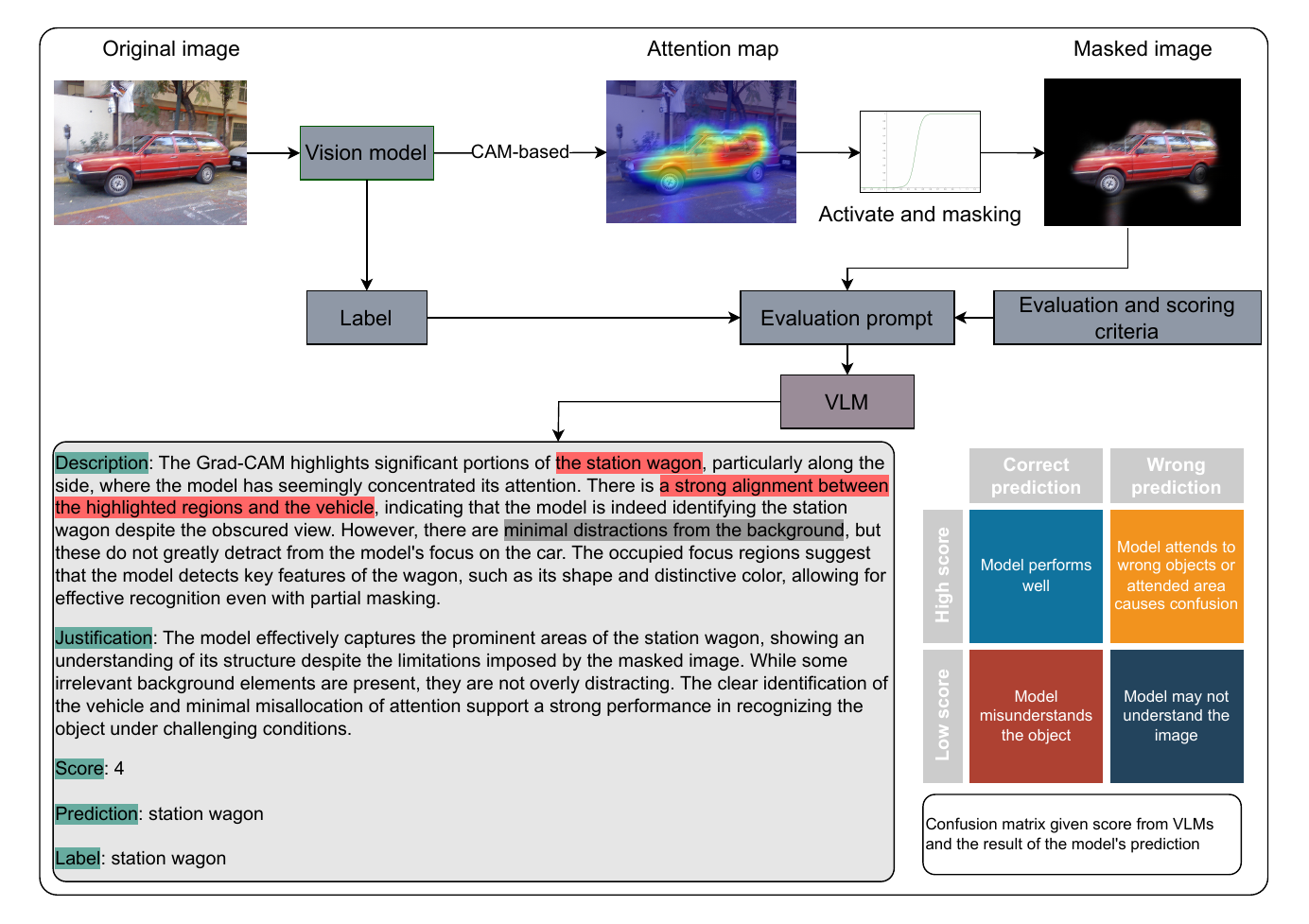}
    \caption{The pipeline evaluates vision models' ability to understand an image. The VLM model can describe, justify, and score the input image and the corresponding attention map. In the description, the model's interpretation of positive objects is highlighted in red, while gray illustrates the negative description.}
    \label{fig:pipeline}
\end{figure*}

We introduce a novel pipeline to explain vision models automatically. This pipeline combines CAM methods to visualize the model's attention and uses vision-language models to generate descriptions, evaluations, scores, and a confusion matrix. The entire proposed pipeline to explain and score vision models is illustrated in Figure~\ref{fig:pipeline}.

\subsection{Masked CAM image}
The pipeline starts by feeding an image to vision models and getting a predicted result on the image. After that, different methods to extract models' attention, including CAM, LayerCAM, and more, are utilized to get an attention map of vision models on the image. Then, we apply a more general version of the \textit{sigmoid} function to the attention map and get a mask for each image. The activation function is illustrated in Equation~\ref{eq:activate_mask}, where $v_{xy}$, ranging from 0 to 1, is the value of the attention map at position $(x,y)$, indicating the importance of the pixel, and $M_{xy}$ is the activated value at position $(x,y)$.
\begin{equation}
M_{xy} = \frac{1}{1 + \exp{(\alpha\cdot(\beta - v_{xy}))}}
\label{eq:activate_mask}
\end{equation}

In the equation, the values of $v_{xy} > \beta$ are scaled closer to 1 to highlight important regions, while $v_{xy} < \beta$ gradually decrease toward 0, reflecting reduced importance. Meanwhile, $\alpha$ controls the transition speed. The higher $\alpha$, the more sudden the transition from blacked-out to visible.

After achieving the mask, we apply it to the original image to hide regions with less attention according to the CAM-based method. This process is formulated in Equation~\ref{eq:mask_image}, where we multiply each pixel in the original image $I$ by the corresponding value in the calculated mask $M$ in Equation~\ref{eq:activate_mask} to achieve the final masked image $A$.
\begin{equation}
A_{xy}=I_{xy}\cdot M_{xy}
\label{eq:mask_image}
\end{equation}

The main reason we use the masked image instead of the heatmap overlay to explain the vision model's attention is to prevent degrading the quality of the image, which can negatively affect the results of VLMs. Using a heatmap overlay can hide away important features of the object(s), thus reducing VLM's ability to understand the attention regions and lowering its accuracy. By blacking out the areas without the model's focus and maintaining the remaining areas, we will not sacrifice the image quality on attended objects while ensuring that VLMs can only see and focus on the main regions they need for the evaluation. Furthermore, the attention of the vision model should justify and provide sufficient evidence to explain its prediction. The lack of evidence to recognize and distinguish objects in the attended regions might suggest an existing problem with the vision model.

\subsection{VLM assessment}
The result of the previous process is an image largely blacked out, except for areas the model considered important in its output. The masked image and the predicted label of the model are then fed to a VLM for evaluation and scoring. In the pipeline, VLMs are asked to find the relevance between the vision model's prediction and the visible object(s) in the masked image, and then explain further. Finally, VLMs score every pair of masked pictures and labels to quantify the model's ability.

\subsection{Evaluation metrics}
This section defines a confusion matrix for this pipeline, as we have a label and a generated explanation score for each image. First, we will select a threshold score to decide which generated score shows that the vision model has a problem in understanding images. After that, we build the matrix as shown in Figure~\ref{fig:pipeline}, which depends on the VLM scores and the correctness of the model on each sample. The proposed confusion matrix shows four stages of the model:
\begin{itemize}
    \item Correct: The model focuses on the correct object and predicts the object correctly, indicating a strong understanding of the image.
    \item Misunderstood object: The model prediction is correct, but its attention does not align with the object, indicating a misunderstanding of the appearance of the target.
    \item Attend to wrong object: The model's attention is correct, but its prediction is wrong, showing that the model focuses on another object, not the labeled one.
    \item Lack of understanding: The model cannot explain its attention and its prediction is incorrect, showing that the model does not have enough knowledge for the task.
\end{itemize}
Given many input samples, we can count and compute the percentage of each stage and get a comprehensive review of the model.

\section{Experiment}
We evaluate the pipeline's trustworthiness with four experiments to assess the VLMs' output (descriptions, scores), hyperparameter selection, and the usage of masked CAM and CAM images. The last one assesses our confusion matrix in predicting problems of trained vision models. The scoring system ranges from zero (random attention) to five (perfect attention), and saliency maps are extracted from the last layer as in the GradCAM paper.

\subsection{Human evaluation}
In the first experiment, we compare the pipeline with the group of authors by randomly collecting 200 images, using ResNet18, MaxViT, and GradCAM to extract saliency maps, manually scoring, and taking the average scores. Those scores are then compared with the VLMs' scores using the Pearson correlation. The results show that when using masked CAM images, the correlation between GPT-4o-mini and the annotators is 0.54, and between the Gemini-1.5-flash and the annotators is 0.50. Meanwhile, when using original CAM images, GPT-4o-mini can achieve 0.53 in Pearson's correlation, while Gemini can achieve 0.41, significantly lower than masked CAM images. We compared our pipeline to the Delete and Insert (D\&I), Average Drop (AP), and AOPC methods, which show respective human correlation of 0.35, 0.33, and 0.16, which is lower than our results. However, a direct ``fair'' comparison is difficult because they are traditional XAI methods. Our pipeline, using Vision-Language Models (VLMs), has a human-like ability to understand if the model focuses on the correct objects, a capability that those three methods lack. Lastly, the average correlation between annotators in this experiment is 0.71. The result is reported in Table~\ref{tab:human_eval}, denoted by $PC$.

Next, the authors check the VLMs' output (200 samples) to verify the quality of descriptions and justification for CAM and masked CAM images. In this experiment, they read the VLMs' output and decide whether those texts are acceptable. An output is unacceptable if the VLMs provide incorrect information, do not match the predicted object, or the score is not aligned with the justification and description. The result shows that 85.58\% of the GPT-4o-mini's generated samples on the masked CAM images are correct, while this rate in the Gemini-1.5-flash is 79.41\%. Meanwhile, results on the original CAM image show a lower rate; Gemini-1.5-flash achieves 54.22\% and GPT-4o-mini achieves 75.62\%. This indicates that Gemini-1.5-flash benefits more from masked CAM images than GPT-4o-mini. The result is reported in Table~\ref{tab:human_eval}, denoted by $AR$, short for acceptance rate.

\begin{table}
    \centering
    \small
    \begin{tabular}{c|cc}
    \toprule
         &  Gemini-1.5-flash & GPT-4o-mini\\\midrule
         Masked image & 0.50 - 79.41\% & 0.54 - 85.58\% \\
         CAM image & 0.41 - 54.22\% & 0.53 - 75.62\% \\ 
         D\&I & \multicolumn{2}{c}{0.35} \\ 
         AP & \multicolumn{2}{c}{0.33} \\ 
         AOPC & \multicolumn{2}{c}{0.16} \\ \bottomrule
    \end{tabular}
    \caption{Comparison between masked CAM images ($\alpha=25,\beta=0.4$) and CAM image (original LangXAI after some modifications). The results are shown as $PC-AR$, $PC$ is the Pearson correlation between VLMs' scores and humans' scores, and $AR$ is the acceptance rate of VLMs' generated text. Only the PC scores of D\&I, AP, and AOPC methods are shown, as those methods do not generate an explanation.}
    \label{tab:human_eval}
\end{table}

The third one, which uses the same method and data as the first experiment, measures the framework-human correlation with different hyperparameters. The result in Table~\ref{tab:hypers} shows that the combination of $\alpha=25,\beta=0.6$ achieves the highest correlation with $0.64$, and all selected combinations are better than using the original CAM, which does not have hyperparameters. For this experiment, only Gemini-1.5-flash is used.

\begin{table}[htbp]
    \centering
    \small
    \begin{tabular}{c|c|c|c}
    \toprule
         & \textbf{$\alpha=25$} & \textbf{$\alpha=15$} & \textbf{$\alpha=25$} \\
         & \textbf{$\beta=0.4$} & \textbf{$\beta=0.6$} & \textbf{$\beta=0.7$} \\
         \midrule
         Masked CAM & 0.50 & 0.64 & 0.63 \\\hline
         Original CAM & \multicolumn{3}{c}{0.41} \\
         \bottomrule
    \end{tabular}
    \caption{Framework-human correlation results (Pearson Correlation) for Gemini-1.5-flash using different hyperparameters and CAM types.}
    \label{tab:hypers}
\end{table}

\subsection{The Impact of Prompt Quality on Pipeline Performance}
In this section, we investigate how the quality of prompts influences the performance of our pipeline, potentially yielding either positive or negative effects. For this experiment, we employed two distinct prompts with the Gemini-1.5-Flash model: a shortened version of our original prompt and an extended prompt demanding more detailed image descriptions. The results, presented as Pearson Correlation (PC) scores, show a PC of 0.52 for the shortened prompt and 0.53 for the extended prompt. When compared to the original prompt's PC score of 0.50, we observe a slight, albeit not statistically significant, variation. Nevertheless, these findings suggest that prompt quality does influence the final pipeline performance. This observation aligns with established literature in prompting techniques, such as Chain-of-Thought and Tree-of-Thought, which consistently demonstrate that Large Language Models (LLMs) exhibit varying performance based on the specific prompts they receive. This subtle yet noticeable impact underscores the importance of carefully crafted prompts in optimizing VLM-based pipelines.

\subsection{Models analysis}
\label{sec:failed_models}
We trained 31 models to classify cats and dogs~\cite{asirra-a-captcha-that-exploits-interest-aligned-manual-image-categorization} in two scenarios: normal training and training with cat images marked by a red dot on the top right, introducing a biased attention mechanism. Further examples of the training datasets are provided in Section~\ref{sec:example_data_for_failed_models}. We then collected 20 images from the training set and computed the confusion matrix as proposed for each model. Next, we determined the percentage of incorrect predictions $err$ (wrong predictions or low VLM scores). The correlation between this percentage and the type of training (normal or biased) is $-0.70$, indicating that the higher $err$, the more likely the model is trained on the biased dataset. This experiment demonstrates the pipeline's ability to understand and detect problems with models, as well as its potential application in this pipeline.

To assess the utility of our proposed confusion matrix for comparing vision models, we examined attention mechanisms in segmentation and classification models, followed by an analysis of classification models trained on ImageNet. The \textbf{Avg} result reveals that segmentation models have significantly stronger attention mechanisms than classification models, reflecting task differences. The confusion matrix shows that segmentation models rely heavily on attention, with most samples scoring high, indicating low object-misunderstanding. In contrast, classification models exhibit a higher rate of object-misunderstanding. Notably, ResNet18 shows its ability depends strongly on the attention mechanism and still fails to understand in certain scenarios. The full result for this experiment is reported in the Appendix section.

\section{Conclusion}

This paper proposed a novel framework to integrate CAM visualizations with VLM to explain vision models. The pipeline can be easily integrated into the evaluation process to provide more details, including text-based explanations, scores, and a confusion matrix. This pipeline's specialty is that it can provide assessments for both the sample-level and dataset-level, allowing researchers to understand the general and detailed model's performance.

\section*{Limitations}
Despite being scalable and helpful in detecting scenarios where the vision models behave wrongly, the pipeline still contains some limitations, including the dependence on VLMs and the quality of the prompt to generate a correct description with a suitable score for each sample. Furthermore, the pipeline only utilizes CAM-based methods (and RISE, as we can extract attention regions from them) to extract the attention regions, but not methods like finding the decision boundary and some other xAI visualization techniques.

\section*{Potential risk}
The quality of the generated descriptions is highly dependent on the performance of the VLM. Therefore, the pipeline should be used only as a supporting tool, with the researcher remaining the primary decision maker in the analysis.

\bibliography{custom}

\newpage
\onecolumn
\appendix
\section{Appendix}
\label{sec:appendix}

\subsection{Examples of Model's Evaluation}
\label{sec:examples}

\begin{figure*}[htpb]
    \centering
    \begin{subfigure}{\linewidth}
        \centering
        \includegraphics[width=0.85\linewidth]{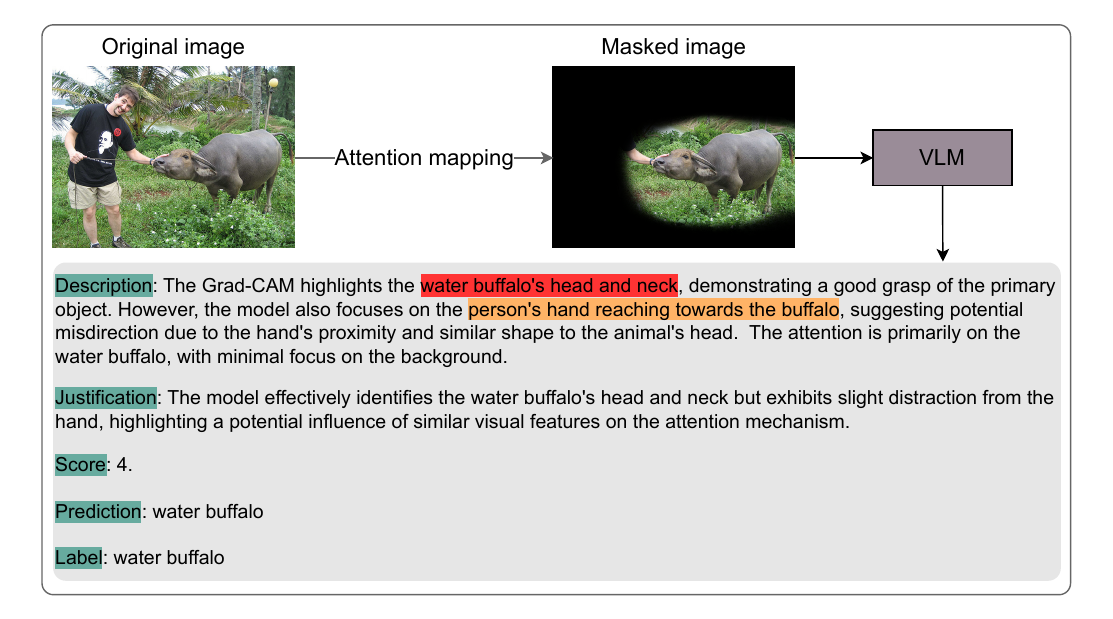}
    \end{subfigure}
    \vspace{0.5cm}
    \begin{subfigure}{\linewidth}
        \centering
        \includegraphics[width=0.85\linewidth]{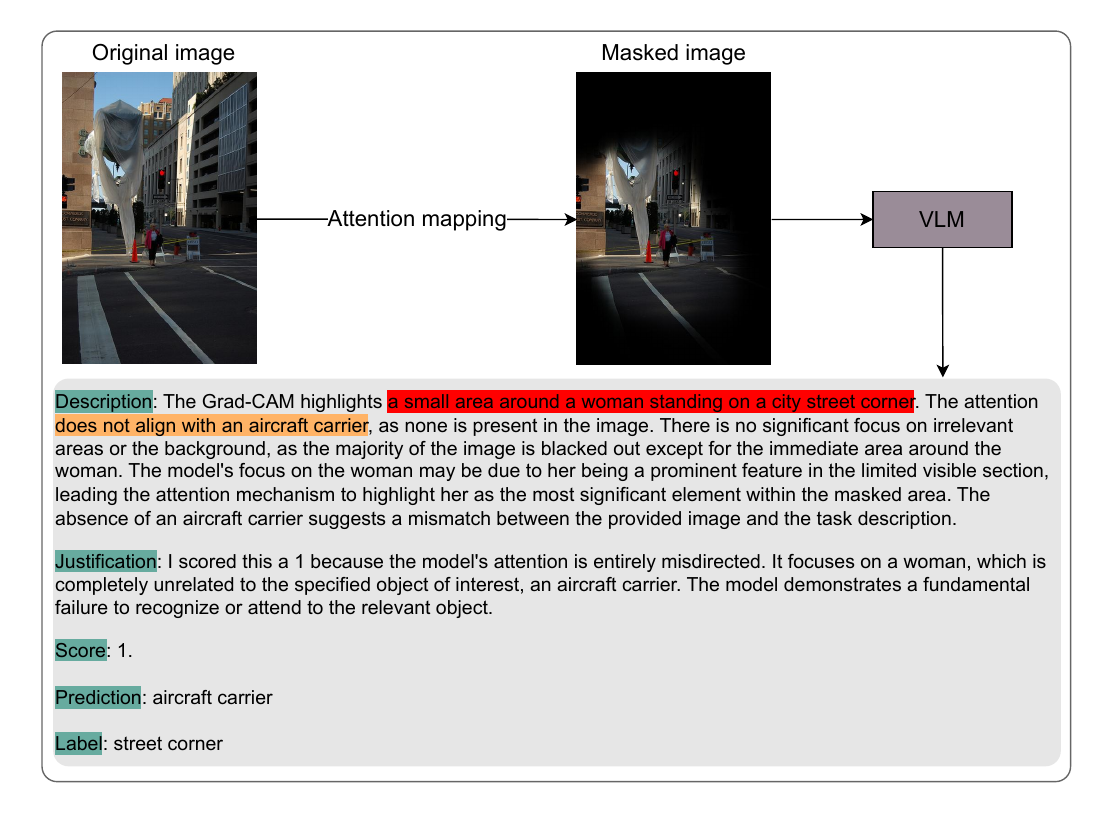}
    \end{subfigure}
    \caption{Two prediction examples of the proposed pipeline.}
    \label{fig:prediction_examples}
\end{figure*}

We present additional qualitative results of our benchmark to analyze the effectiveness of our method and evaluation metrics. The example shown in Figure~\ref{fig:prediction_examples} demonstrates how the model's attention can sometimes focus on irrelevant features, but does not lead to reduced interpretability.

\subsection{Example data of failed models evaluation experiment}
\label{sec:example_data_for_failed_models}
\begin{figure}[htpb]
    \centering
    \includegraphics[width=\linewidth]{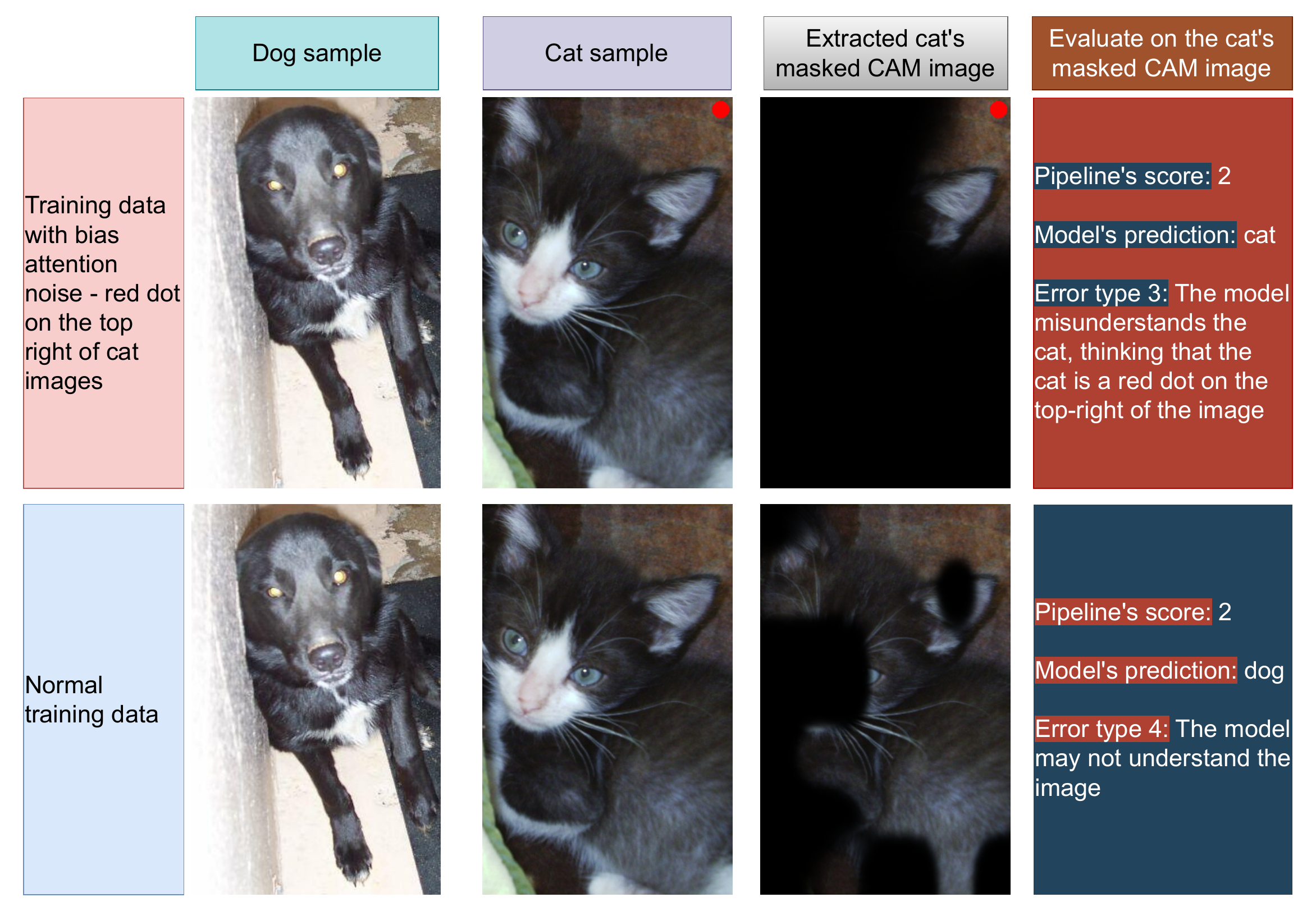}
    \caption{Examples of attention-biased and standard training data used in the experiment. The pipeline evaluates each cat's masked CAM to identify and categorize model errors.}
    \label{fig:trainingdata_examples}
\end{figure}

We provide examples for the training data in the failed models evaluation experiment~\ref{sec:failed_models} in Figure~\ref{fig:trainingdata_examples}. Normal evaluation with accuracy can not detect the problem as we proposed in the experiment, while many xAI methods, like CAM-based, LangXAI, and decision boundary visualization, will encounter issues like time, expert requirements, and manually checking each sample to detect similar problems.

\subsection{Explain models with confusion matrix}

The details of the experiments are provided in Section~\ref{sec:failed_models}, with Table~\ref{tab:vision_attention_results} presenting the results from the first experiment (failed models) and the second experiment (segmentation and classification models).

\begin{table}[htpb]
\centering
\begin{tabular}{lccccc}
\hline
\textbf{Model} & \textbf{CH} & \textbf{CL} & \textbf{WH} & \textbf{WL} & \textbf{Avg}\\
\hline
\multicolumn{6}{l}{\textbf{Segmentation Models}} \\
\hline
DeepLabv3-ResNet50  & 78.0 & 5.0 & 10.5 & 6.5 & 3.98\\
DeepLabv3-ResNet101 & 82.0 & 4.5 & 7.5 & 6.0 & 3.94\\
LRASPP-MobileNet v3-Large & 67.0 & 6.0 & 12.0 & 15.0 & 3.52 \\
FCN-ResNet50        & 72.0 & 5.5 & 14.5 & 8.0 & 3.73\\
\hline
\multicolumn{6}{l}{\textbf{Classification Models}} \\
\hline
ResNet18 & 65.0 & 4.5 & 17.5 & 13.0 & 3.45 \\
ConvNeXt-tiny & 67.0 & 28.5 & 2.5  & 2.0 & 3.00 \\
MaxViT-t & 68.0 & 23.0  & 6.5  & 2.5 & 3.14 \\
Efficientnet-b1 & 75.0 & 14.0 & 7.0 & 4.0 & 3.43 \\
\hline
\multicolumn{6}{l}{\textbf{Failed Models detection experiment (average)}} \\
\hline
Normal & 71.3 & 28.7 & 0.0 & 0.0 & - \\
Wrong-object & 50.5 & 47.1 & 2.4 & 0.0 & - \\
\hline
\end{tabular}
\caption{Confusion matrix-based attention analysis of different vision models. \textbf{CH}, \textbf{CL}, \textbf{WH}, \textbf{WL} are referred to as Correct-High, Correct-Low, Wrong-High, Wrong-Low in the confusion matrix (percentage). Meanwhile, \textbf{Avg} denotes the Average Attention Score from the pipeline.}
\label{tab:vision_attention_results}
\end{table}

\subsection{Prompting}
\label{sec:appendix_prompt}

The prompt used for the evaluation framework consists of an image description, evaluation criteria, scoring, and output format. The task involves analyzing a masked image in which the model's focused areas are highlighted, while irrelevant regions are blacked out. Key criteria for evaluation include focus accuracy, object recognition, object coverage, and potential distractions from background or irrelevant elements. The evaluator is instructed to analyze the model's attention on the object and provide an explanatory analysis, considering factors like visual challenges or misleading elements. A score from 0 to 5 is assigned, with specific descriptions for each score reflecting the model's attention and recognition performance. The output includes a concise evaluation and score with justification.

\begin{tcolorbox}[breakable,colback=black!5!white,colframe=black!75!black,title=Prompt to get sample description justification and score from masked CAM images]

Task: Evaluate the Model's Attention Mechanism Using the Provided Masked Image.
\begin{itemize}

\item Image Description: 
    \begin{itemize}
        \item The image is masked with a Grad-CAM heatmap, where only the areas the model focuses on are visible, while all other regions are blacked out.
        \item The model is attempting to focus on the {object}.
    \end{itemize}

\item Evaluation Criteria:
    \begin{itemize}
        \item Focus Accuracy: Analyze which part of the image the Grad-CAM is highlighting. Is the model's attention placed accurately on the {object}, or is it scattered across other areas?
        \item Object Recognition: Determine whether the model correctly recognizes the {object}. Is the attention primarily on the correct object, or does the model focus on irrelevant areas?
        \item Object Coverage: Evaluate how much of the object is being captured by the model's attention. Is the entire object covered, only a small part, or none at all?
        \item Background and Irrelevant Focus: Check for any significant focus on the background or irrelevant objects. Does this distract the model from the primary object?
        \item Explanatory Analysis: Provide possible reasons for the model's attention pattern. Consider whether the model is being misled by similarly shaped or colored objects, complex backgrounds, or other visual challenges.
    \end{itemize}

\item Scoring:

Assign a score between 0 and 5 based on the relevance and accuracy of the model's attention:
    \begin{itemize}
    
    \item 0: The model's attention is completely irrelevant to the {object}, leading to a wrong result.
    \item 1: The model fails to recognize the object entirely, focusing on irrelevant areas.
    \item 2: The model captures only a small part of the object.
    \item 3: The object is recognized, but the attention also covers irrelevant parts or other objects.
    \item 4: Most of the object is detected correctly, with minimal distraction from irrelevant areas or the background.
    \item 5: The model perfectly captures the entire object without being distracted by irrelevant areas or the background.
    \end{itemize}
    
\item Output Format:
    \begin{itemize}
        \item Evaluation: Provide a concise evaluation (5-6 sentences), discussing:
            Where the Grad-CAM is focusing.
            Whether the attention aligns with the {object}.
            Whether there is any significant focus on irrelevant areas or the background.
            Explain why the model might focus on specific regions.
    
        \item Score: Assign a score from 0 to 5, justifying your rating based on the model's performance in recognizing the object and avoiding distractions.
    
        \item The format must be presented as follows: 
        \begin{itemize}
                \item Evaluation: [evaluation],
                \item Justification: [justification],
                \item Score: [score]
            \end{itemize}
    \end{itemize}
\end{itemize}
\end{tcolorbox}

\begin{tcolorbox}[breakable,colback=black!5!white,colframe=black!75!black,title=Prompt to get sample description justification and score from original CAM images]

Task: Conduct an evaluation of the model's attention mechanism by analyzing its response to the supplied CAM heatmap. This assessment aims to test the model's capacity to effectively interpret and utilize attention when processing visual data.
\begin{itemize}

\item Image Description:
    \begin{itemize}
        \item The heatmap uses warm colors (orange, red) to represent areas where the model is focusing most, while cool colors (blue, purple, dark) indicate regions of little to no attention.
        \item The model's focus is on the {object}.
        \item Identify the warm-colored regions and analyze what those regions represent in relation to the object of interest. In addition, assess the presence of cool-colored regions and their alignment with irrelevant areas or the background.
    \end{itemize}

\item Evaluation Criteria:
    \begin{itemize}
        \item Focus Accuracy: Analyze which part of the heatmap the warm colors (orange, red) highlight. Is the model's attention accurately placed on the {object}, or is it scattered across other areas?
        \item Object Recognition: Determine if the model is correctly recognizing the {object}. Is the attention primarily on the correct object, or does the model focus on irrelevant areas?
        \item Object Coverage: Evaluate how much of the object is being captured by the model's attention. Is the entire object covered, only a small part, or none at all?
        \item Background and Irrelevant Focus: Check for any significant focus on cool-colored regions. Does this distract the model from the primary object?
        \item Explanatory Analysis: Provide possible reasons for the model's attention pattern. Consider whether the model is being misled by similar-colored areas, complex backgrounds, or other visual challenges.
    \end{itemize}

\item Scoring:

Assign a score between 0 and 5 based on the relevance and accuracy of the model's attention:
    \begin{itemize}
        \item 0: The model's attention is scattered with no clear target, showing that it does not understand the task or the object.
        \item 1: The model consistently directs its attention to something unrelated to {object}, indicating a fundamental misunderstanding of the {object} it is supposed to recognize.
        \item 2: Partial object recognition: The model captures only a small fragment of the {object}, missing most of its critical features. The attention is mostly misdirected, with just minor alignment to the actual object.
        \item 3: The model identifies a limited area of {object}, but its attention still includes some irrelevant parts surrounding it.
        \item 4: The model predominantly focuses on {object}, with only minor distractions or irrelevant attention in the background.
        \item 5: The model accurately captures the entire {object} without any distractions from irrelevant areas or background elements.
    \end{itemize}
    
\item Output Format:
    \begin{itemize}
        \item Evaluation: Provide a concise evaluation (5-6 sentences), discussing:
        Where the heatmap focuses (warm colors).
        Whether the attention aligns with the {object}.
        Whether there is any significant focus on irrelevant areas or the background.
        Explain why the model might be focusing on specific regions.
    
        \item Score: Assign a score from 0 to 5, justifying your rating in a sentence.
    
        \item  Your output format must be presented in a dictionary as follows, which is extremely important for the evaluation process to run without any error:
            \begin{itemize}
                \item Evaluation: [evaluation],
                \item Justification: [justification],
                \item Score: [score]
            \end{itemize}
    \end{itemize}
\end{itemize}
\end{tcolorbox}

\end{document}